\documentclass[10pt,letterpaper]{article}

\usepackage{cogsci}

\cogscifinalcopy 

\usepackage{pslatex}
\usepackage{apacite}
\usepackage{float} 

\usepackage{multirow}
\usepackage{times}
\usepackage{latexsym}
\usepackage{amsmath}
\usepackage{amsfonts}
\usepackage{multirow}
\usepackage{booktabs,arydshln}
\usepackage{graphicx}
\usepackage{algorithm}
\usepackage{algorithmic}
\usepackage{tabu}
\usepackage{xcolor}
\usepackage{setspace}

\title{Navigating Brain Language Representations: A Comparative Analysis of Neural Language Models and Psychologically Plausible Models}
 
\author{Yunhao Zhang$^{1,2}$, Shaonan Wang$^{1,2}$, Xinyi Dong$^{3}$, Jiajun Yu$^{4}$,  Chengqing Zong$^{1,2}$ \\
  ${}^1$State Key Laboratory of Multimodal Artificial Intelligence Systems, Institute of Automation, CAS, Beijing, China\\
  ${}^2$School of Artificial Intelligence, University of Chinese Academy of Sciences, Beijing, China\\
  ${}^3$State Key Laboratory of Cognitive Neuroscience and Learning, Beijing Normal University\\
  ${}^4$College of Information and Electrical Engineering, China Agricultural University\\
  zhangyunhao2021@ia.ac.cn; \{shaonan.wang, cqzong\}@nlpr.ia.ac.cn; \\
  202121061097@mail.bnu.edu.cn;  2017307070311@cau.edu.cn\\
}

\begin{document}

\maketitle

\begin{abstract}
Neural language models, particularly large-scale ones, have been consistently proven to be most effective in predicting brain neural activity across a range of studies. However, previous research overlooked the comparison of these models with psychologically plausible ones. Moreover, evaluations were reliant on limited, single-modality, and English cognitive datasets. To address these questions, we conducted an analysis comparing encoding performance of various neural language models and psychologically plausible models. Our study utilized extensive multi-modal cognitive datasets, examining bilingual word and discourse levels. Surprisingly, our findings revealed that psychologically plausible models outperformed neural language models across diverse contexts, encompassing different modalities such as fMRI and eye-tracking, and spanning languages from English to Chinese. Among psychologically plausible models, the one incorporating embodied information emerged as particularly exceptional. This model demonstrated superior performance at both word and discourse levels, exhibiting robust prediction of brain activation across numerous regions in both English and Chinese.

\textbf{Keywords:} 
Neural Language Models, Psychologically Plausible Models, Neural Encoding
\end{abstract}

\section{Introduction}

Neural language models, particularly large ones, exhibit remarkable effectiveness in diverse downstream tasks and exceptional language understanding abilities. This success has spurred the development of studies utilizing neural language models to investigate how the human brain processes languages. Previous research utilizes representations generated by distinct models to predict brain activation \cite{mitchell2008predicting, huth2016natural}, assuming that the closer the representation aligns with the brain's semantic information, the better it captures brain activation.

Recent research has compellingly demonstrated that neural language models excel in predicting brain neural activities \cite{schrimpf2021neural}. However, existing studies have overlooked the comparison with simple yet psychologically plausible models, which possess mathematically transparent computational mechanisms and do not necessitate extensive hyperparameter tuning processes. Furthermore, these studies often rely on a single modality of brain activity data from small cognitive datasets to evaluate models. These datasets typically involve stimuli from a single paradigm and a single language unit (e.g., word-level). Moreover, the focus of these investigations has primarily been on Germanic languages, particularly English, while Tibetan languages like Chinese remain largely unexplored.

Therefore, the effectiveness of neural language models in predicting brain activations and their superiority over other models, as well as their ability to capture the nuances of human language processing, remains uncertain. Comprehensive studies are essential to understand the intricate relationship between artificial intelligence and the brain's cognitive functions.

To bridge this gap, we conduct a comparative analysis, evaluating the encoding performance of diverse neural language models (NLMs) and psychologically plausible models (PPMs) on eight multi-modal cognitive datasets in Chinese and English, encompassing both word and discourse levels. To investigate spatial difference of model encoding performance, we perform a fine-grained analysis at both the region of interest (ROI) and voxelwise levels using fMRI data. Our findings reveal that: (i) Simple PPMs outperform NLMs across various contexts, encompassing diverse modalities like fMRI and eye-tracking and spanning languages from English to Chinese. (ii) PPMs integrating embodied and network-topological information excel in word-level encoding, with embodied models outperforming others at the discourse level. PPMs incorporating local-statistical information are particularly adept at fitting eye-tracking patterns. (iii) The shallow layers of NLMs excel in word-level brain activation, whereas middle layers are better at capturing discourse-level activation. (iv) The brain cortex encoding map reveals unique correlations between different models and various brain regions, implying distinct and exclusive information encoding within different models.

\begin{figure*}[t]
\centering 
\includegraphics[width=0.95\textwidth]{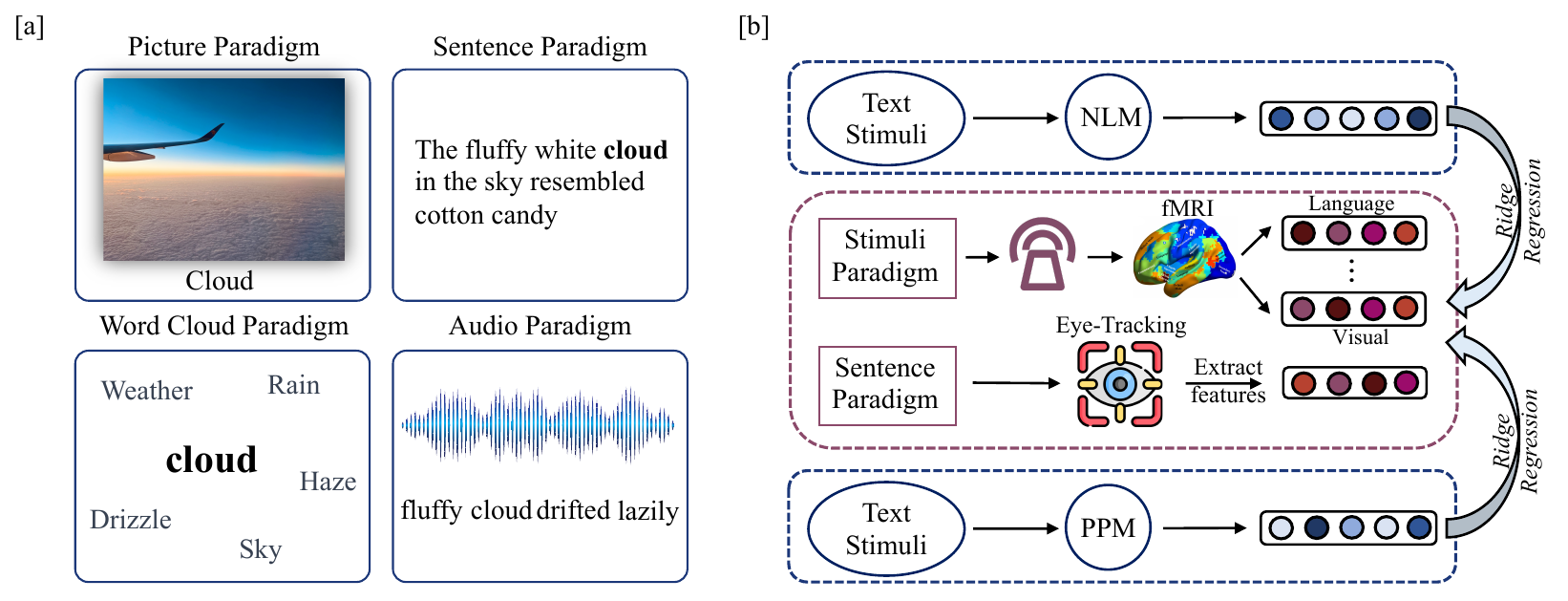}
\caption{[a] Stimuli examples from four paradigms: picture, sentence, and word cloud paradigms are employed in word-level fMRI, while the audio paradigm is used in discourse-level fMRI. Sentence paradigm is utilized in eye-tracking data. [b] Neural encoding method using representations generated by neural language models and psychologically plausible models.}\label{fig:main_pro}
\vspace{-3mm}
\end{figure*}

\section{Related Work}
\subsection{Evaluating Neural Language Models with Cognitive Data}
Previous research has attempted to identify models that more effectively capture brain activation. \citeA{abnar2017experiential} evaluates seven NLMs, encompassing distributional, dependency-based, and one experiential model, in predicting neural responses to 60 nouns presented by \citeA{mitchell2008predicting}. \citeA{beinborn2019robust} evaluates the ability of ELMo to predict brain responses on four small fMRI datasets. These studies demonstrate NLMs' effective prediction of brain activation. \citeA{hollenstein2019cognival} evaluates six NLMs consisting of four context-independent models and two context-aware models on relatively more cognitive datasets. Moreover, \citeA{schrimpf2021neural} compares multiple state-of-art models on relatively small sentence-level datasets, from no more than 9 participants. These studies conclude that context-aware models perform best in encoding English cognitive data.

However, recent work has demonstrated that learning and evaluation goals commonly utilized with NLMs, like masked or next-word prediction, are inconsistent with the brain's language understanding mechanisms \cite{pasquiou2022neural,antonello2022predictive}. Therefore, NLMs may not be the optimal model for capturing brain activity. Moreover, cognitive datasets used in these studies are either too small or only contain stimuli from single paradigm and single language unit. Furthermore, these studies have predominantly centered around Germanic languages, particularly English, while Tibetan languages, such as Chinese, remain largely unexplored. Investigating the relationship between model representations and brain activation in diverse languages can enrich our comprehension of brain language representations \cite{wang2024computational}.

\subsection{Psychologically Plausible Models}
In psycholinguistics, three distinct types of computational principles are considered as candidates for the semantic representational algorithm in the human brain. 

The first type is local-statistical system, such as simple co-occurrence, whose variations can be detected by humans during language acquisition \cite{saffran2001acquisition, conway2005modality}. Several studies have found that co-occurrence information in language corpora predicts various human semantic phenomena \cite{roelke2018novel,hofmann2018simple,frank2017word}, indicating that co-occurrence information likely plays a pivotal role in shaping human semantic organization.

The second type is global-network-topological system. In network sciences, language can be conceptualized as a complex network, with words as nodes and their correlations as edges, displaying rich topological properties. Numerous studies have shown that humans implicitly infer these topological properties during various structural learning tasks, including motor sequence learning \cite{lynn2020abstract}, object relation learning \cite{garvert2017map}, visual event segmentation \cite{schapiro2013neural}, and picture naming \cite{fu2023different}.

The third type is embodied-based system. It is motivated by grounded semantics, where experiential information from diverse modality-specific systems, such as visual, motor, and social, is re-encoded into semantic representations stored in memory. Several studies have found that embodied-based system partly reflects conceptual knowledge in human brain \cite{damasio1989time, glenberg1997memory, binder2011neurobiology, fernandino2022decoding}.

The aforementioned studies primarily investigate the relationship between PPMs and brain’s neural representations at word level. It remains unclear to what extent PPMs encode the brain's neural representations of sentences and discourse, along with behavioral signals like eye-tracking. Furthermore, these studies do not incorporate NLMs for a comprehensive comparison and analysis.

\setlength{\tabcolsep}{1.5mm}
\begin{table*}[htbp]
	\centering
	\renewcommand\arraystretch{1}
    \begin{tabular}{|c|c|c|c|c|c|c|} \cline{1-7}
       \textbf{Language}& \textbf{Modality} & \textbf{Source} & \textbf{Paradigm} & \textbf{Subject} &\textbf{Unit} & \textbf{Tokens} \\ \cline{1-7}
       \multirow{5}{*}{English} &Word fMRI &  \citeA{pereira2018toward} & Picture & 15 &Word & 180 \\ \cline{2-7}
       &Word fMRI &  \citeA{pereira2018toward} & Text & 15& Word & 180 \\ \cline{2-7}
       &Word fMRI &  \citeA{pereira2018toward} & Word Cloud & 15& Word & 180 \\ \cline{2-7}
       &Discourse fMRI &  \citeA{zhang2020connecting} & Audio & 19& Discourse & 47,356 \\ \cline{2-7}
       &Eye-tracking &  \citeA{hollenstein2018zuco} & Text & 12& Sentence & 36,767 \\ \cline{1-7}
       \multirow{3}{*}{Chinese} & Word fMRI &  \citeA{wang2022fmri} & Picture & 11& Word & 672\\ \cline{2-7}
       & Discourse fMRI & \citeA{wang2022synchronized} & Audio & 12& Discourse & 52,269 \\ \cline{2-7}
       & Eye-tracking &  \citeA{zhang2022database} & Text & 1718& Sentence & 170,331 \\ \cline{1-7}
    \end{tabular}
    \caption{Details of the cognitive datasets used in our experiments.}
    \label{tab:cogdata}
     \vspace{-3mm}
\end{table*}

\section{Cognitive Datasets}
We introduce the sources of English and Chinese cognitive datasets (See Table \ref{tab:cogdata} for details.). For word-level fMRI data, we preprocess it using fMRIPrep \cite{esteban2019fmriprep} and conduct first-level analysis to obtain t-value images representing neural activation for each word. Discourse-level fMRI data is preprocessed using the Human Connectome Project (HCP) pipeline. To align word representations with fMRI data, the representations are convolved with the hemodynamic response function (HRF) and down-sampled to the discourse-fMRI sampling rate. In eye-tracking data, four key word-level features are extracted, encompassing the entire reading process. These features, categorized into total reading time (TRT) and gaze duration (GD), number of fixations (nFixations), and first fixation duration (FFD), capture various aspects of processing.

\section{Method}

In this section, we demonstrate how the representations produced by neural language models and psychologically plausible models are used to encode cognitive data.

\subsection{Neural Language Models}
We adopt six typical neural language models that can be divided into two groups: one is context-independent models including GloVe \cite{pennington2014glove} and Word2Vec \cite{mikolov2013efficient}, where GloVe is a count-based method that performs a dimensionality reduction on the co-occurrence matrix, and Word2Vec employs a shallow neural network to map words from a large corpus into continuous vector spaces. The other is context-aware models including GPT2\textsubscript{\textnormal{ori}}, GPT2\textsubscript{\textnormal{med}}, BERT\textsubscript{\textnormal{base}} and BERT\textsubscript{\textnormal{lar}}, where GPT2 is an auto-regressive language model, trained to predict the next token based on preceding text, and BERT is an auto-encoder language model, trained bidirectionally to predict masked tokens.

To obtain word embeddings, we train GloVe and Word2Vec on large-scale corpora, the Xinhua News corpus (19.7 GB)\footnote{http://www.xinhuanet.com/whxw.htm} for Chinese and the Wikipedia corpus (13 GB)\footnote{https://dumps.wikimedia.org/enwiki/latest} for English, using identical model parameters, including the Skip-Gram architecture, a negative number of 15 for Word2Vec, a window width of 2, and embedding dimensions of 300. GPT2\textsubscript{\textnormal{ori}} and BERT\textsubscript{\textnormal{base}} have 12 hidden layers, GPT2\textsubscript{\textnormal{med}} and BERT\textsubscript{\textnormal{large}} have 24 hidden layers, and we take word representations from each hidden layer. We derive each word's representation by averaging sub-word embeddings for discourse stimuli and adopt \citeA{chersoni2021decoding} and \citeA{zhang2023comprehensive}'s methods to extract word representations for word stimuli.

\subsection{Psychologically Plausible Models}
We developed three psychologically plausible models, each based on a unique computational principle for semantic representation in the human brain. \\
\textbf{Local-Statistical Model (LSM)} Following previous studies \cite{fu2023different, frank2017word}, we construct the local-statistical model by calculating a co-occurrence matrix weighted by Positive Pointwise Mutual Information and reducing it with Singular Value Decomposition. Finally, we obtain a 300-dimensional representation for each word. \\
\textbf{Network-Topological Model (NTM)} Inspired by previous studies \cite{newman2017second}, we first calculate the one-order similarity coefficients (e.g., cosine similarity) among words. Then, we construct a graph, where each node is a word and each edge is cosine similarity coefficients. Finally, we use random walks and dimensionality reduction techniques to generate network-topological word representations. \\
\textbf{Embodied-Based Model (EBM)} Word representations with six semantic dimensions were obtained from a prior study \cite{wang2023large}, which represent various aspects of word meanings, such as vision, motor, socialness, emotion, time, and space, in alignment with neural processing systems \cite{binder2016toward}. We obtain a 6-dimensional representation encompassing both sensory-motor and abstract semantic information for each word.

\subsection{Encoding Brain Activation}
As shown in Figure \ref{fig:main_pro}[b], to map representations yielded by different models (NLMs and PPMs) to brain activations, for each cognitive dataset, we train encoding models to predict fMRI or eye-tracking signals from text stimuli for each subject. Specifically, we follow K-fold (K=10) cross-validation. All data samples from K-1 folds were utilized for training, while the model is tested on the remaining fold. Following previous works \cite{oota2022neural}, we employ sklearn's ridge regression, 10-fold cross-validation, mean squared error loss function. Moreover, we perform a group-level paired t-test with false discovery rate (FDR) correction to assess the significance of comparison results.

\begin{figure*}[t]
\centering 
\setlength{\textfloatsep}{2pt}
\setlength{\intextsep}{2pt}
\setlength{\abovecaptionskip}{2pt}
\includegraphics[width=0.85\textwidth]{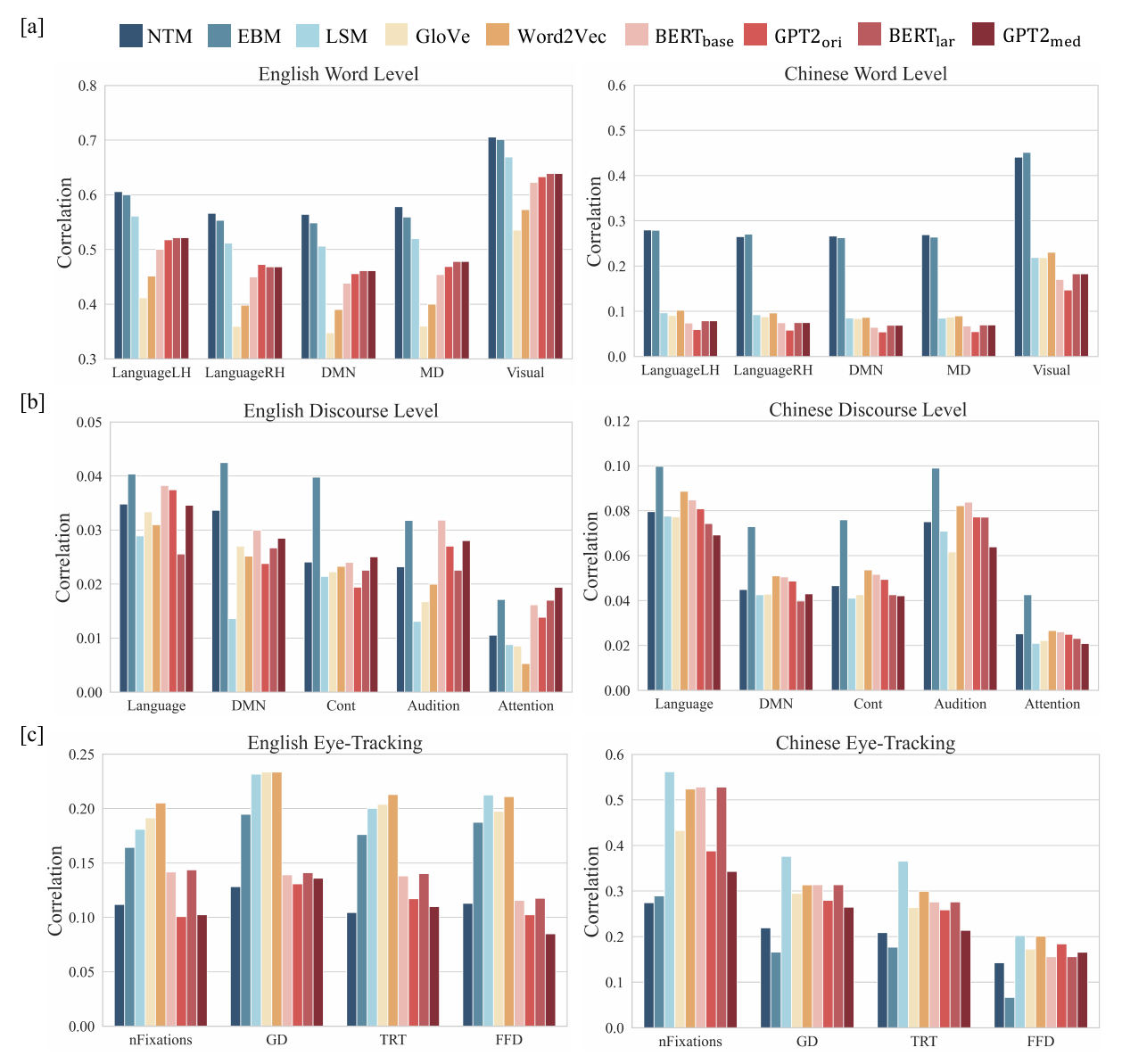}
\caption{Pearson correlation coefficients between predicted and true values were computed for English and Chinese in word-level fMRI, discourse-level fMRI, and eye-tracking data using both NLMs and PPMs. Results are averaged across all subjects. To facilitate comparison, we average results from the picture paradigm, sentence paradigm, and word cloud paradigm to obtain the English word-level results. As for context-aware models, we select the layer with the best performance.}\label{fig:main_result}
\end{figure*}

\section{Result and Discussion}
\begin{figure*}[t]
\centering 
\setlength{\textfloatsep}{2pt}
\setlength{\intextsep}{2pt}
\setlength{\abovecaptionskip}{2pt}
\includegraphics[width=0.9\textwidth]{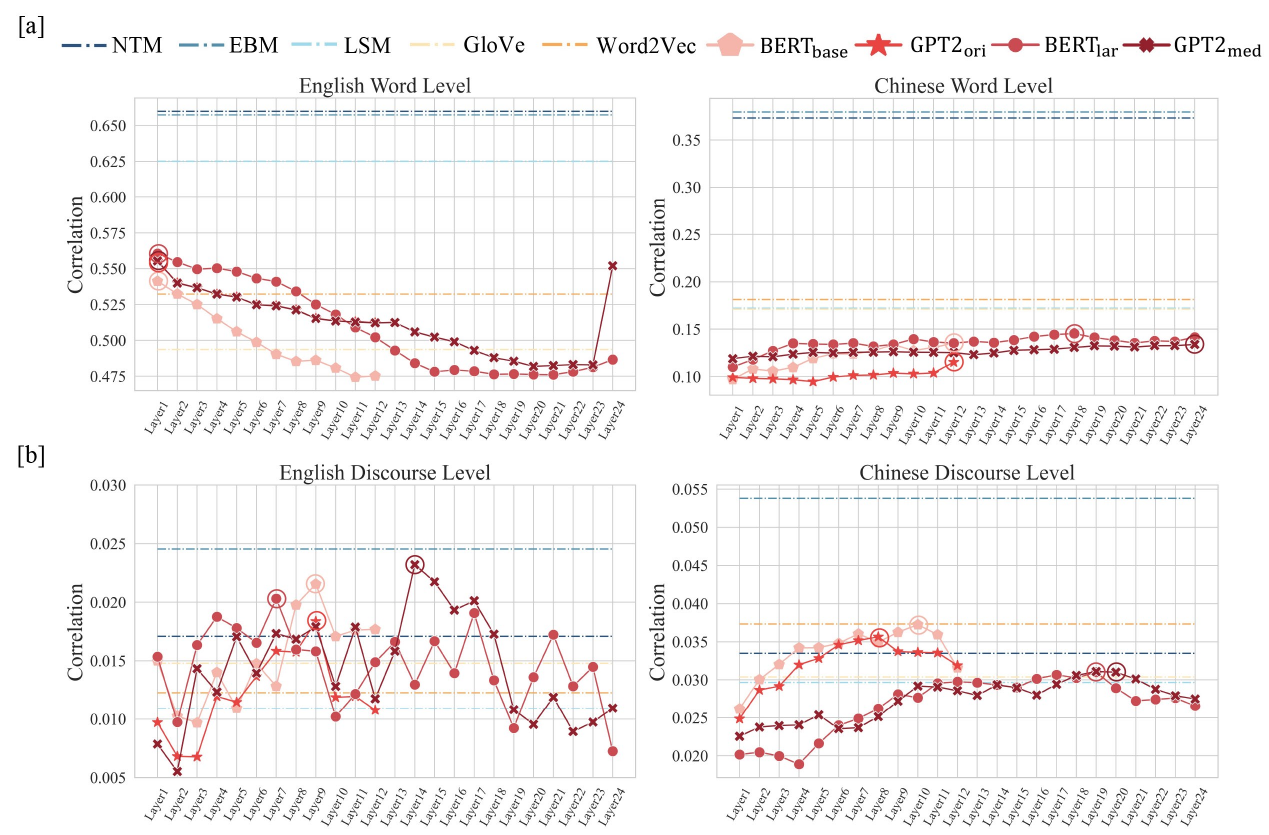}
\caption{Mean Pearson correlation coefficients were calculated for each layer of context-aware models in English and Chinese word-level and discourse-level fMRI data across the entire brain. A circle marks the layer with the best encoding performance for each model.}\label{fig:main_layer}
\vspace{-3mm}
\end{figure*}

Figure \ref{fig:main_result} illustrates the mean encoding performance of NLMs and PPMs across subjects for selected brain networks in Chinese and English, covering word-level fMRI, discourse-level fMRI and eye-tracking data. Figure \ref{fig:main_layer} displays the mean performance of each layer within context-aware models across whole brain for both Chinese and English in word-level and discourse-level fMRI. Figure \ref{fig:analysis_discourse} shows distribution of top-performing model for each voxel across whole brain.

\subsection{Comparison between Neural Language Models and Psychologically Plausible Models}
Figure \ref{fig:main_result} illustrates that at the word level, PPMs significantly outperform NLMs across various brain networks and languages like English and Chinese ($p<0.05$, FDR corrected). At the discourse level, PPMs show superior average performance across most brain networks ($p<0.05$, FDR corrected). However, in brain networks with less emphasis on language processing (Audition and Attention), NLMs exhibit comparable performance to PPMs in English. In eye-tracking, PPMs have comparable performance to NLMs on average ($p>0.05$, FDR corrected). Moreover, Figure \ref{fig:main_layer} indicates that at the word level, PPMs outperform each layer of context-aware models ($p<0.05$, FDR corrected). In the discourse level, PPMs on average outperform NLMs across the entire brain. In summary, PPMs prove to be more effective in predicting brain activation than NLMs.

In word level, the better performance in PPMs supports popular research views that network-topological and embodied properties explain certain aspects of the brain's conceptual representation mechanism, respectively \cite{fu2023different, fernandino2022decoding}. The lower performance in NLMs may be attributed to the learning and evaluation goals frequently used with NLMs, such as masked or next-word prediction, which are inconsistent with the brain's language comprehension mechanisms. Consequently, NLM-generated representations lack similarity to human brain semantic information. Furthermore, the performance of context-aware models is relatively close to PPMs at the discourse level compared to the word level. It's probably because context-aware models can obtain the specific meaning of a word within its context. Therefore, they encode brain activation for complex language unit containing context, such as discourse, more effectively.

\subsection{Comparison within Psychologically Plausible Models}
\textbf{Word Level} Both EBM and NTM obtain best performance in predicting word-level brain activation in English and Chinese. And in Figure \ref{fig:main_result}[a], LSM performs similarly to NTM and EBM in the English paradigm but lags behind in the Chinese paradigm. Compared to English, Chinese has more intricate grammar and text structure \cite{tang2021differences}. Therefore, the simple local-statistical model like LSM struggles to capture word-level brain activation from Chinese text corpus. 



\begin{figure*}[tbp]
\centering 
\setlength{\textfloatsep}{2pt}
\setlength{\intextsep}{2pt}
\setlength{\abovecaptionskip}{2pt}
\includegraphics[width=0.8\textwidth]{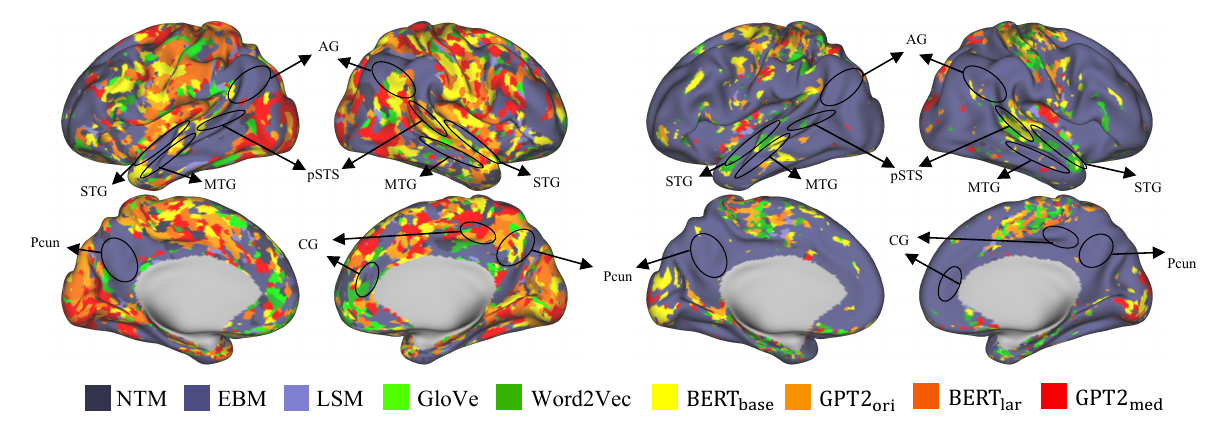}
\caption{Distribution of top-performing model for each voxel across the entire brain in English (left) and Chinese (right) discourse-level fMRI. ROIs related to semantic processing are marked.}\label{fig:analysis_discourse}
\vspace{-3mm}
\end{figure*}

\noindent \textbf{Discourse Level} Figure \ref{fig:main_result}[b] illustrates EBM's superior encoding performance in various brain networks. This can be attributed to the situational nature of discourse-level tasks \cite{speelman1990representation}, which require higher-order cognitive functions like imaginative scenario construction and benefit from more sensory-motor information involvement. Moreover, LSM obtains the lowest performance in PPMs, indicating that local-statistical representations can not effectively capture brain activation during comprehending content at discourse level.


\noindent \textbf{Eye-Tracking} As shown in Figure \ref{fig:main_result}[c], LSM outperforms other PPMs on nearly all eye-tracking features, in line with psychological research indicating that local statistical patterns, such as co-occurrence, can predict reading behavior, which is measured using eye-tracking data \cite{mcdonald2003eye}. Moreover, NTM and EBM achieve relatively low encoding performance, suggesting that effectively modeling brain activation may not achieve the same effect for behavioral signals.

\subsection{Comparison within Neural Language Models} 
\textbf{Word Level} As shown in Figure \ref{fig:main_result}[a], we observe that context-aware models have better encoding performance than that of context-independent models in the English paradigm, while context-independent models excel in context-aware models in the Chinese paradigm. In psycholinguistics, the holistic view suggests that due to the prevalence of homographic and homophonic morphemes in Chinese, holistic processing of compound words is more efficient \cite{packard1999lexical}. And neuroimaging studies have highlighted the brain's reliance on holistic word-level processing during Chinese semantic learning \cite{tsang2022erp}. Therefore, BERT and GPT2 models in Chinese acquire semantic information through character-level training, diverging from human brain processing. Yet Word2Vec in Chinese learns word-level representations from the large-scale corpus, which have better encoding performance than context-aware models. 

As shown in Figure \ref{fig:main_layer}[a], in the Chinese paradigm, context-aware models show consistent encoding performance across various layer depths, whereas in the English paradigm, they exhibit significant variation across various layer depths. This result suggests that, in English context-aware models, more semantic information akin to the human brain is present in shallow layers, whereas in Chinese context-aware models, it is evenly distributed across layers. Furthermore, GPT2\textsubscript{\textnormal{med}} and BERT\textsubscript{\textnormal{large}} has better performance than GPT2\textsubscript{\textnormal{origin}} and BERT\textsubscript{\textnormal{base}}. This result suggests that context-aware models with more parameters capture additional word-level semantic information akin to the human brain. 

\noindent \textbf{Discourse Level} 
As observed in Figure \ref{fig:main_result}[b], context-aware models outperform context-independent models in English, while the context-independent model (Word2Vec) excels context-aware models in Chinese, in line with word-level findings.

As shown in Fig \ref{fig:main_layer}[b], context-aware models show a rise followed by a decline in performance as the layer rises, which is similar on English and Chinese datasets. However, the shallow layers of context-aware models have higher performance than the deep layers on English word-level datasets. These language-unit differences between layers suggest that, in context-aware models, shallow layers encode simple linguistic units, while middle layers capture richer semantic information related to more complex linguistic units \cite{zhang2024mulcogbench}.

\noindent \textbf{Eye-Tracking} It can be noticed from Figure \ref{fig:main_result}[c] that in English, context-independent models outperform context-aware models, contrary to fMRI results. It indicates that effectively modeling brain activation may not yield the same impact on behavioral signals. 

Moreover, NLMs accurately predict TRT and FFD in English, while they predict TRT better than FFD in Chinese, which may indicate that Chinese models capture the information of late-stage semantic integration better than early processing stages of lexical access.

\subsection{Encoding Performance on Each Voxel} 
To thoroughly explore encoding performance across various models, we label each voxel using the model exhibiting the best performance for that voxel. As illustrated in Figure \ref{fig:analysis_discourse}, EBM consistently achieves superior encoding performance in nearly all ROIs related to semantic processing, highlighting its ability to capture brain activation during language comprehension. Furthermore, we observe that distinct models capture activation from different brain regions. For example, GPT2\textsubscript{\textnormal{med}} captures activation in regions such as the lateral occipital cortex (LocG), insular gyrus (INS), and superior parietal lobule (SPL), while BERT\textsubscript{\textnormal{base}} captures activation in regions including the angular gyrus (AG), middle frontal gyrus (MFG), posterior superior temporal sulcus (pSTS), middle temporal gyrus (MTG), and precentral gyrus (PrG). The cortical encoding map indicates unique correlations between specific models and various brain regions, suggesting distinct and exclusive information encoding within these models.


\section{Conclusion}
Analyzing multi-modal cognitive data across diverse contexts and languages, our study shows that psychologically plausible models outperform neural language models in brain encoding performance, challenging current claims of neural models' excellence in predicting brain activation during language processing. 
Moreover, our results replicate existing findings regarding the encoding performance of neural language models with English cognitive data, as well as discover the prominent contribution of embodied and network-topological information to brain activation prediction in both English and Chinese.

Our findings offer valuable insights for selecting computational models in diverse cognitive tasks, shedding light on language processing across languages and contexts, guiding future research.

\section{Acknowledgements}
We would like to thank the anonymous reviewers for their valuable comments. This research was supported by grants from the National Natural Science Foundation of China to S.W. (62036001) and S.W. (the STI2030-Major Project, grant number: 2021ZD0204105).

\bibliographystyle{apacite}

\setlength{\bibleftmargin}{.125in}
\setlength{\bibindent}{-\bibleftmargin}

\bibliography{CogSci_Template}

\end{document}